\documentclass{article}

\usepackage{amsmath,amsfonts,bm}
\allowdisplaybreaks

\newcommand{\graph}{\mathcal{G}}

\newcommand{\MLP}{\mathrm{MLP}}
\newcommand{\MPN}{\mathrm{MPN}}

\newcommand{\set}[1]{\{ #1 \}}

\def\eqref#1{equation~\ref{#1}}

\def\1{\bm{1}}

\def\vmu{{\bm{\mu}}}

\def\vSigma{{\bm{\Sigma}}}

\def\ve{{\bm{e}}}

\def\vg{{\bm{g}}}
\def\vh{{\bm{h}}}

\def\vp{{\bm{p}}}

\def\vz{{\bm{z}}}

\DeclareMathAlphabet{\mathsfit}{\encodingdefault}{\sfdefault}{m}{sl}
\SetMathAlphabet{\mathsfit}{bold}{\encodingdefault}{\sfdefault}{bx}{n}

\def\gD{{\mathcal{D}}}
\def\gE{{\mathcal{E}}}

\def\gN{{\mathcal{N}}}

\def\gQ{{\mathcal{Q}}}

\def\gS{{\mathcal{S}}}

\def\gV{{\mathcal{V}}}

\newcommand{\softmax}{\mathrm{softmax}}

\usepackage{hyperref}       
\usepackage{url}            
\usepackage{booktabs}       
\usepackage{algorithm}
\usepackage{algorithmic}

\usepackage{amsfonts}       
\usepackage{nicefrac}       
\usepackage{microtype}      
\usepackage{enumitem}
\usepackage{multirow}
\usepackage{graphicx}
\usepackage{cleveref}
\usepackage{subcaption}

\usepackage{hyperref}

\usepackage[accepted]{icml2020}

\icmltitlerunning{Multi-Objective Molecule Generation using Interpretable Substructures}

\begin{document}

\twocolumn[
\icmltitle{Multi-Objective Molecule Generation using Interpretable Substructures}



\icmlsetsymbol{equal}{*}

\begin{icmlauthorlist}
\icmlauthor{Wengong Jin}{mit}
\icmlauthor{Regina Barzilay}{mit}
\icmlauthor{Tommi Jaakkola}{mit}
\end{icmlauthorlist}

\icmlaffiliation{mit}{MIT CSAIL}

\icmlcorrespondingauthor{Wengong Jin}{wengong@casil.mit.edu}

\icmlkeywords{Machine Learning}

\vskip 0.3in
]



\printAffiliationsAndNotice{}  

\begin{abstract}
Drug discovery aims to find novel compounds with specified chemical property profiles. In terms of generative modeling, the goal is to learn to sample molecules in the intersection of multiple property constraints. This task becomes increasingly challenging when there are many property constraints. We propose to offset this complexity by composing molecules from a vocabulary of substructures that we call molecular rationales. These rationales are identified from molecules as substructures that are likely responsible for each property of interest. We then learn to expand rationales into a full molecule using graph generative models. Our final generative model composes molecules as mixtures of multiple rationale completions, and this mixture is fine-tuned to preserve the properties of interest. We evaluate our model on various drug design tasks and demonstrate significant improvements over state-of-the-art baselines in terms of accuracy, diversity, and novelty of generated compounds.
\end{abstract}
\section{Introduction}

The key challenge in drug discovery is to find molecules that satisfy multiple constraints, from potency, safety, to desired metabolic profiles. Optimizing these constraints simultaneously is challenging for existing computational models. The primary difficulty lies in the lack of training instances of molecules that conform to all the constraints. For example, for this reason, \citet{jin2019multi} reports over 60\% performance loss when moving beyond the single-constraint setting.  

\begin{figure*}[t]
    \centering
    \includegraphics[width=\textwidth]{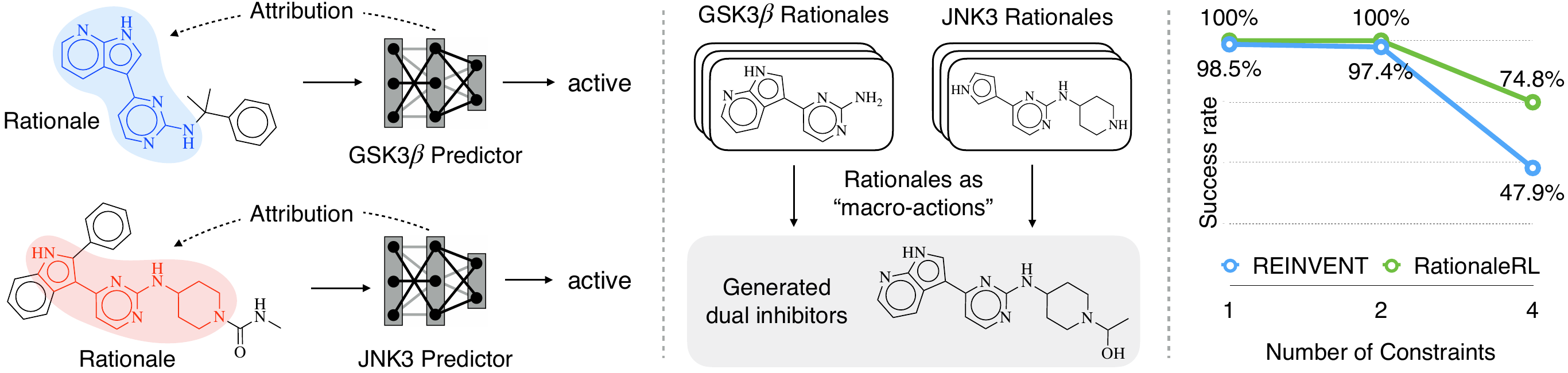}
    \caption{Illustration of RationaleRL. \emph{Left}: To generate a dual inhibitor against biological targets GSK3$\beta$ and JNK3, our model first identifies rationale substructures $\gS$ for each property. Note that rationales are not provided as domain knowledge. \emph{Middle}: The model learns to compose multiple rationales $\gS$ into a complete molecule $\graph$. \emph{Right:} Our method achieves much higher success rate than the current state-of-the-art molecule design method REINVENT~\cite{olivecrona2017molecular}) under four property constraints.}
    \label{fig:motivation}
\end{figure*}

In this paper, we propose a novel approach to multi-property molecular optimization. Our strategy is inspired by fragment-based drug discovery~\citep{murray2009rise} often followed by medicinal chemists. The idea is to start with
substructures (e.g., functional groups or later pieces) that drive specific properties of interest, and then combine these building blocks into a target molecule. To automate this process, our model has to learn two complementary tasks: (1) identification of the building blocks that we call rationales, and (2) assembling multiple rationales together into a fully formed target molecule. In contrast to competing methods, our generative model does not build molecules from scratch, but instead assembles them from automatically extracted rationales already implicated for specific properties (see Figure~\ref{fig:motivation}). 

We implement this idea using a generative model of molecules where the rationale choices play the role of latent variables. Specifically, a molecular graph $\graph$ is generated from underlying rationale sets $\gS$ according to: 
\begin{equation}
    P(\graph) = \sum_\gS\nolimits P(\graph | \gS) P(\gS)
\end{equation}
As ground truth rationales (e.g., functional groups or subgraphs) are not provided, the model has to extract candidate rationales from molecules with the help of a property predictor. 
We formulate this task as a discrete optimization problem efficiently solved by Monte Carlo tree search.
Our rationale conditioned graph generator, $P(\graph | \gS)$, is initially trained on a large collection of real molecules so that it is capable of expanding any subgraph into a full molecule. The  mixture model is then fine-tuned using reinforcement learning to ensure that the generated molecules preserve all the properties of interest. This training paradigm enables us to realize molecules that satisfy multiple constraints without observing any such instances in the training set. 

The proposed model is evaluated on molecule design tasks under different combinations of property constraints.\footnote{https://github.com/wengong-jin/multiobj-rationale} 
Our baselines include state-of-the-art molecule generation methods~\citep{olivecrona2017molecular,you2018graph}. 
Across all tasks, our model achieve state-of-the art results in terms of accuracy, novelty and diversity of generated compounds. In particular, we outperform the best baseline with 38\% absolute improvement in the task with three property constraints.
We further provide ablation studies to validate the benefit of our architecture in the low-resource scenario. Finally, we show that identified rationales are chemically meaningful in a toxicity prediction task~\citep{sushko2012toxalerts}.

\section{Related Work}
\label{sec:relwork}
\textbf{Reinforcement Learning } One of the prevailing paradigms for drug design is reinforcement learning (RL)~\citep{you2018graph,olivecrona2017molecular,popova2018deep}, which seeks to maximize the expected reward defined as the sum of predicted property scores using the property predictors. Their approach learns a distribution $P(\graph)$ (a neural network) for generating molecules. Ideally, the model should achieve high success rate in generating molecules that meet all the constraints, while maintaining the diversity of $P(\graph)$.

The main challenge of RL is the reward sparsity, especially when there are multiple competing constraints. For illustration, we test a state-of-the-art reinforcement learning method~\citep{olivecrona2017molecular} under four property constraints: biological activity to GSK3$\beta$, JNK3, drug-likeness and synthetic accessibility~\citep{li2018multi}. As shown in Figure~\ref{fig:motivation}, initially the success rate and diversity is high when given only one of the constraints, but they decrease dramatically when all the property constraints are added. 
The reason of this failure is that the property predictor (i.e., reward function) remains black-box and the model has limited understanding of why certain molecules are desirable.

Our framework offsets this complexity by understanding property landscape through rationales. At a high level, the rationales are analogous to \emph{options}~\citep{sutton1999between,stolle2002learning}, which are macro-actions leading the agent faster to its goal. The rationales are automatically discovered from molecules with labeled properties. 

\textbf{Molecule Generation } 
Previous work have adopted various approaches for generating molecules under specific property constraints.
Roughly speaking, existing methods can be divided along two axes --- representation and optimization. 
On the representation side, they either operate on SMILES strings~\citep{gomez2016automatic,segler2017generating,kang2018conditional} or directly on molecular graphs~\citep{simonovsky2018graphvae,jin2018junction,samanta2018nevae,liu2018constrained,de2018molgan,ma2018constrained,seff2019discrete}.
On the optimization side, the task has been formulated as reinforcement learning~\citep{guimaraes2017objective,olivecrona2017molecular,popova2018deep,you2018graph,zhou2018optimization}, continuous optimization in the latent space learned by variational autoencoders \citep{gomez2016automatic,kusner2017grammar,dai2018syntax-directed,jin2018junction,kajino2018molecular,liu2018constrained}, or graph-to-graph translation~\citep{jin2018learning}.
In contrast to existing approaches, our model focuses on the multi-objective setting of the problem and offers a different formulation for molecule generation based on rationales.


\textbf{Interpretability } 
Our rationale based generative model seeks to provide transparency~\citep{doshi2017towards} for molecular design. The choice of rationales $P(\gS)$ is visible to users and can be easily controlled by human experts. Prior work on interpretability primarily focuses on finding rationales (i.e., explanations) of model predictions in image and text classification~\citep{lei2016rationalizing,ribeiro2016should,sundararajan2017axiomatic} and molecule property prediction~\citep{mccloskey2019using,ying2019gnnexplainer,lee2019functional}. In contrast, our model uses rationales as building blocks for molecule generation.

\section{Proposed Approach: RationaleRL}

\begin{figure*}[t]
    \centering
    \includegraphics[width=\textwidth]{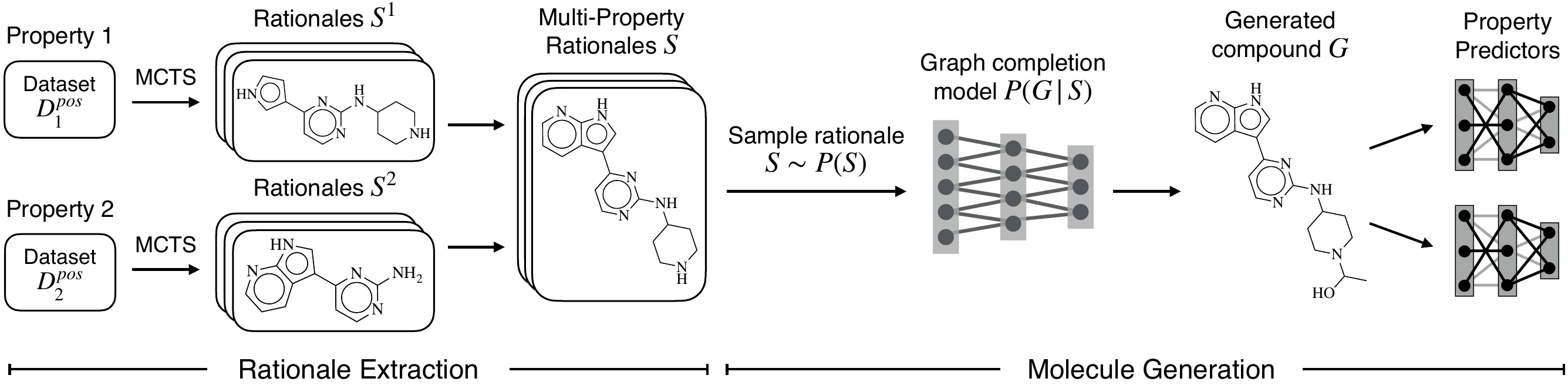}
    \caption{Overview of our approach. We first construct rationales for each individual property and then combine them as multi-property rationales. The method learns a graph completion model $P(\graph | \gS)$ and rationale distribution $P(\gS)$ in order to generate positive molecules.}
    \label{fig:overview}
\end{figure*}

Molecules are represented as graphs $\graph = (\gV , \gE)$ with atoms $\gV$ as nodes and bonds $\gE$ as edges. The goal of drug discovery is to find novel compounds satisfying given property constraints (e.g., drug-likeness, binding affinity, etc.). Without loss of generality, we assume the property constraints to be of the following form:
\begin{equation}
\begin{split}
    \text{Find molecules} \quad & \graph  \\
    \text{Subject to} \quad & r_i(\graph) \geq \delta_i; \quad i=1,\cdots,M
\end{split}
\label{eq:design}
\end{equation}
For each property $i$, the property score $r_i(\graph) \in [0, 1]$ of molecule $\graph$ must be higher than threshold $\delta_i \in [0, 1]$. 
A molecule $\graph$ is called \emph{positive} to property $i$ if $r_i(\graph) \geq \delta_i$ and \emph{negative} otherwise.

Following previous work~\citep{olivecrona2017molecular,popova2018deep}, $r_i(\graph)$ is output of property prediction models (e.g., random forests) which effectively approximate empirical measurements. The prediction model is trained over a set of molecules with labeled properties gathered from real experimental data. The property predictor is then \emph{fixed} throughout the rest of the training process.

\textbf{Overview  } Our model generates molecules by first sampling a rationale $\gS$ from the vocabulary $V_{\gS}^{[M]}$, and then completing it into a molecule $\graph$. The generative model is defined as
\begin{equation}
P(\graph) = \sum_{\gS \in V_{\gS}^{[M]}}\nolimits P(\gS) P(\graph | \gS) 
\end{equation}
As shown in Figure~\ref{fig:overview}, our model consists of three modules:
\begin{itemize}[leftmargin=*,topsep=0pt,itemsep=0pt]
    \item \textbf{Rationale Extraction}: Construct rationale vocabulary $V_{\gS}^i$ each individual property $i$ and combines these rationales for multiple properties $V_\gS^{[M]}$ (see \S\ref{sec:rationale}).
    
    \item \textbf{Graph Completion} $P(\graph | \gS)$: Generate molecules $\graph$ using multi-property rationales $S^{[M]} \in V_\gS^{[M]}$. The model is first pre-trained on natural compounds and then fine-tuned to generate molecules satisfying multiple constraints (see \S\ref{sec:completion} for its architecture and \S\ref{sec:training} for fine-tuning).
    
    \item \textbf{Rationale Distribution} $P(\gS)$: The rationale distribution $P(\gS)$ is learned based on the properties of complete molecules $\graph$ generated from $P(\graph | \gS)$. A rationale $\gS$ is sampled more frequently if it is more likely to be expanded into a positive molecule $\graph$ (see \S\ref{sec:training}).
\end{itemize}

\subsection{Rationale Extraction from Predictive Models}
\label{sec:rationale}

\textbf{Single-property Rationale } 
We define a rationale $\gS^i$ for a single property $i$ as a subgraph of some molecule $\graph$ which causes $\graph$ to be active (see Figure \ref{fig:motivation}). To be specific, let $V_{\gS}^i$ be the vocabulary of such rationales for property $i$. Each rationale $\gS^i \in V_{\gS}^i$ should satisfy the following two criteria to be considered as a rationale:
\begin{enumerate}[leftmargin=*,topsep=0pt,itemsep=0pt]
    \item The size of $\gS^i$ should be small (less than 20 atoms).
    \item Its predicted property score $r_i(\gS^i) \geq \delta_i$.
\end{enumerate}

For a single property $i$, we propose to extract its rationales from a set of \emph{positive} molecules $\gD_i^{pos}$ used to train the property predictor. For each molecule $\graph_i^{pos} \in \gD_i^{pos}$, we find a rationale subgraph with high predicted property and small size ($N_s=20$): 
\begin{equation}
\begin{split}
    \text{Find subgraph} \quad & \gS^i \subset \graph_i^{pos} \\
    \text{Subject to} \quad & r_i(\gS^i) \geq \delta_i, \\
    & |\gS^i| \leq N_s \text{ and $\gS^i$ is connected}
\end{split}
\label{eq:find-rationale}
\end{equation}
Solving the above problem is challenging because rationale $\gS^i$ is discrete and the potential number of subgraphs grows exponentially to the size of $\graph_i^{pos}$.
To limit the search space, we have added an additional constraint that $\gS^i$ has to be a connected subgraph.\footnote{This assumption is valid in many cases. For instance, rationales for toxicity (i.e., toxicophores) are connected subgraphs in most cases \citep{sushko2012toxalerts}.} 
In this case, we can find a rationale $\gS^i$ by iteratively removing some peripheral bonds while maintaining its property. Therefore, the key is learning to prune the molecule. 

\begin{figure}[t]
    \centering
    \includegraphics[width=0.47\textwidth]{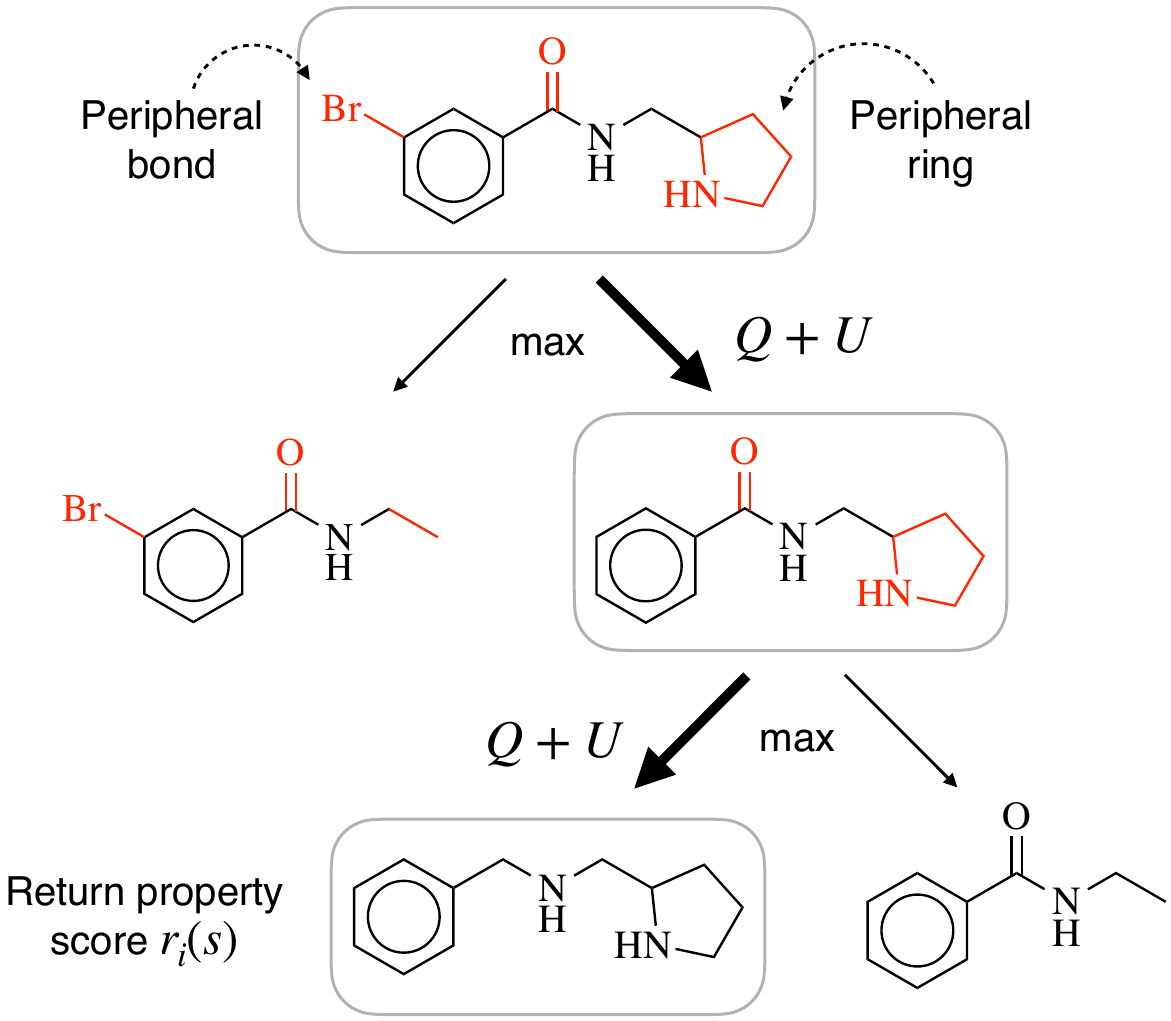}
    \caption{Illustration of Monte Carlo tree search for molecules. Peripheral bonds and rings are highlighted in red. In the forward pass, the model deletes a peripheral bond or ring from each state which has maximum $Q+U$ value (see Eq.(\ref{eq:ufunc})). In the backward pass, the model updates the statistics of each state.}
    \label{fig:mcts}
\end{figure}

This search problem can be efficiently solved by Monte Carlo Tree Search (MCTS)~\citep{silver2017mastering}. The root of the search tree is $\graph^{pos}$ and each state $s$ in the search tree is a subgraph derived from a sequence of bond deletions. To ensure that each subgraph is chemically valid and stays connected, we only allow deletion of one peripheral non-aromatic bond or one peripheral ring from each state. As shown in Figure~\ref{fig:mcts}, a bond or a ring $a$ is called peripheral if $\graph^{pos}$ stays connected after deleting $a$.

During search process, each state $s$ in the search tree contains edges $(s, a)$ for all legal deletions $a$. Following \citet{silver2017mastering}, each edge $(s,a)$ stores the following statistics:
\begin{itemize}[leftmargin=*,topsep=0pt,itemsep=0pt]
    \item $N(s,a)$ is the visit count of deletion $a$, which is used for exploration-exploitation tradeoff in the search process. 
    \item $W(s,a)$ is total action value which indicates how likely the deletion $a$ will lead to a good rationale.
    \item $Q(s,a)$ is the mean action value: $W(s,a) \;/\; N(s,a)$
    \item $R(s,a)=r_i(s')$ is the predicted property score of the new subgraph $s'$ derived from deleting $a$ from $s$.
\end{itemize}

Guided by these statistics, MCTS searches for rationales in multiple iterations. Each iteration consists of two phases:
\begin{enumerate}[leftmargin=*,topsep=0pt,itemsep=0pt]
    \item \emph{Forward pass}: Select a path $s_0,\cdots,s_L$ from the root $s_0$ to a leaf state $s_L$ with less than $N$ atoms and evaluate its property score $r_i(s_L)$. At each state $s_k$, an deletion $a_k$ is selected according to the statistics in the search tree:
    \begin{eqnarray}
        a_k &=& \arg\max_a Q(s_k,a) + U(s_k,a) \\
        U(s_k,a) &=& c_{\mathrm{puct}} R(s_k,a) \frac{\sqrt{\sum_b N(s_k,b)}}{1 + N(s_k,a)} \label{eq:ufunc}
    \end{eqnarray}
    where $c_{puct}$ determines the level of exploration. This search strategy is a variant of the PUCT algorithm~\citep{rosin2011multi}. It initially prefers to explore deletions with high $R(s,a)$ and low visit count, but asympotically prefers deletions that are likely to lead to good rationales.
    \item \emph{Backward pass}: The edge statistics are updated for each state $s_k$. Specifically, $N(s_k,a_k) \leftarrow N(s_k,a_k) + 1$ and $W(s_k,a_k) \leftarrow W(s_k,a_k) + r_i(s_L)$.
\end{enumerate}
In the end, we collect all the leaf states $s$ with $r_i(s) \geq \delta_i$ and add them to the rationale vocabulary $V_{\gS}^i$.

\begin{figure}[t]
    \centering
    \includegraphics[width=0.44\textwidth]{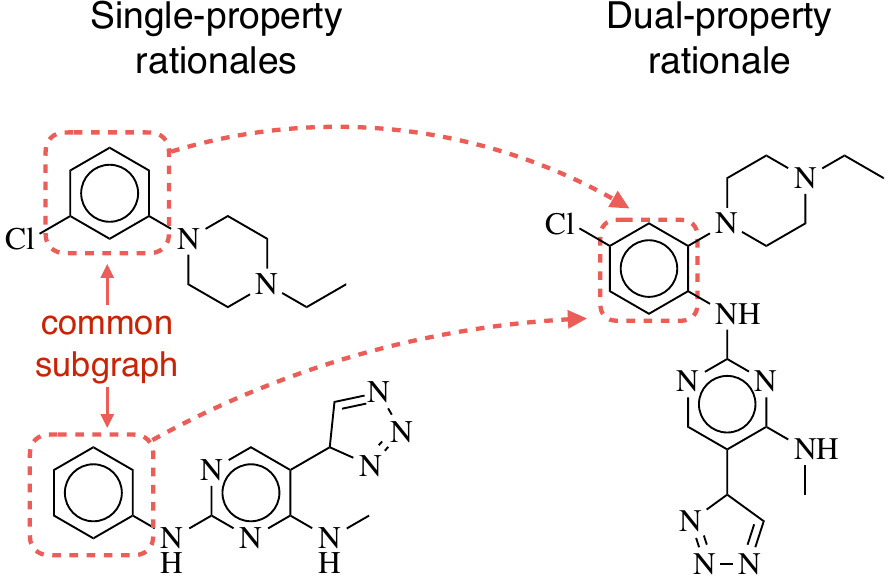}
    \caption{Illustration of multi-property rationale construction. Given two single-property rationales, we first find their maximum common substructure (MCS). If their MCS is not empty, we superpose one rationale on another so that their MCS coincides.}
    \label{fig:merge}
\end{figure}

\textbf{Multi-property Rationale } 
For a set of $M$ properties, we can similarly define its rationale $\gS^{[M]}$ by imposing $M$ property constraints at the same time, namely 
$$\forall i: r_i(\gS^{[M]}) \geq \delta_i, i=1,\cdots,M$$
In principle, we can apply MCTS to extract rationales from molecules that satisfy all the property constraints.
However, in many cases there are no such molecules available. To this end, we propose to construct multi-property rationales from single-property rationales extracted by MCTS. Specifically, each multi-property rationale $\gS^{[M]}$ is merged from single-property rationales $\gS^1, \cdots, \gS^M$.
We merge two rationales $\gS^i$ and $\gS^j$ by first finding their maximum common substructure (MCS) and then superposing $\gS^i$ on $\gS^j$ so that their MCS coincides (see Figure~\ref{fig:merge}).\footnote{The MCS of two (or multiple) rationales is computed using RDKit~\cite{landrum2006rdkit}.}
This gives us a set of candidate rationales:
\begin{equation}
    C_\gS^M = \bigcup_{(\gS^1, \cdots, \gS^M)} \mathrm{MERGE}(\gS^1, \cdots, \gS^M) 
\end{equation}
where $(\gS^1, \cdots, \gS^M) \in V_{\gS}^1 \times \cdots \times V_{\gS}^M$. Note that the output of $\mathrm{MERGE}$ is a set as there are multiple ways of superposing two rationales. Finally, the vocabulary of multi-property rationales is the subset of $C_\gS^M$ which satisfies all the property constraints:
\begin{equation}
    V_\gS^{[M]} = \set{\gS \in C_\gS^M \;|\; r_i(\gS^{[M]}) \geq \delta_i, \forall i}
\end{equation}
For notational convenience, we will denote both single and multi-property rationales as $\gS$ from now on.

\subsection{Graph Completion}
\label{sec:completion}
This module is a variational autoencoder which completes a full molecule $\graph$ given a rationale $\gS$. Since each rationale $\gS$ can be realized into many different molecules, we introduce a latent variable $\vz$ to generate diverse outputs:
\begin{equation}
P(\graph | \gS) = \int_{\vz} P(\graph | \gS, \vz)  P(\vz) d\vz
\end{equation}
where $P(\vz)$ is the prior distribution. Different from standard graph generation, our graph decoder must generate graphs that contain subgraph $\gS$. Our VAE architecture is adapted from existing atom-by-atom generative models~\citep{you2018graphrnn,liu2018constrained} to incorporate the subgraph constraint.
For completeness, we present our architecture here:

\textbf{Encoder } Our encoder is a message passing network (MPN) which learns the approximate posterior $Q(\vz | \graph, \gS)$ for variational inference. Let $\ve(a_u)$ be the embedding of atom $u$ with atom type $a_u$, and $\ve(b_{uv})$ be the embedding of bond $(u,v)$ with bond type $b_{uv}$. The MPN computes atom representations $\set{\vh_v | v \in \graph}$.
\begin{equation}
    \set{\vh_v} = \MPN_e\left(\graph, \set{\ve(a_u)}, \set{\ve(b_{uv})} \right)
\end{equation}
For simplicity, we denote the MPN encoding process as $\MPN(\cdot)$, which is detailed in the appendix. The atom vectors are aggregated to represent $\graph$ as a single vector $\vh_\graph = \sum_v\vh_v$. Finally, we sample latent vector $\vz_\graph$ from $Q(\vz | \graph, \gS)$ with mean $\vmu(\vh_\graph)$ and log variance $\vSigma(\vh_\graph)$:
\begin{equation}
    \vz_\graph = \vmu(\vh_\graph) + \exp(\vSigma(\vh_\graph)) \cdot \epsilon; \quad \epsilon \sim \gN(\mathbf{0}, \mathbf{I})
\end{equation}

\textbf{Decoder } The decoder generates molecule $\graph$ according to its breadth-first order. In each step, the model generates a new atom and all its connecting edges.
During generation, we maintain a queue $\gQ$ that contains frontier nodes in the graph who still have neighbors to be generated.
Let $\graph_t$ be the partial graph generated till step $t$.
To ensure $\graph$ contains $\gS$ as subgraph, we set the initial state of $\graph_1 = \gS$ and put all the peripheral atoms of $\gS$ to the queue $\gQ$ (only peripheral atoms are needed due to the rationale extraction algorithm). 

In $t^\mathrm{th}$ generation step, the decoder first runs a MPN over current graph $\graph_t$ to compute atom representations $\vh_v^{(t)}$:
\begin{equation}
    \set{\vh_v^{(t)}} = \MPN_d\left(\graph_t, \set{\ve(a_u)}, \set{\ve(b_{uv})} \right)
\end{equation}
The current graph $\graph_t$ is represented as the sum of its atom vectors $\vh_{\graph_t} = \sum_{v \in \graph_t} \vh_v^{(t)}$. 
Suppose the first atom in $\gQ$ is $v_t$. The decoder decides to expand $\graph_t$ in three steps:

\begin{enumerate}[leftmargin=*,topsep=0pt,itemsep=0pt]
\item Predict whether there will be a new atom attached to $v_t$: 
\begin{equation}
    \vp_t = \mathrm{sigmoid}(\MLP(\vh_{v_t}^{(t)}, \vh_{\graph_t}, \vz_\graph)) 
\end{equation}
where $\MLP(\cdot, \cdot, \cdot)$ is a ReLU network whose input is a concatenation of multiple vectors.

\item If $\vp_t < 0.5$, discard $v_t$ and move on to the next node in $\gQ$. Stop generation if $\gQ$ is empty. Otherwise, create a new atom $u_t$ and predict its atom type:
\begin{equation}
    \vp_{u_t} = \softmax(\MLP(\vh_{v_t}^{(t)}, \vh_{\graph_t}, \vz_\graph))
\end{equation}

\item Predict the bond type between $u_t$ and other frontier nodes in $\gQ=\set{q_1,\cdots,q_n}$ ($q_1=v_t$). Since atoms are generated in breadth-first order, there are no bonds between $u_t$ and atoms not in $\gQ$. 
\end{enumerate}

To fully capture edge dependencies, we predict the bonds between $u_t$ and atoms in $\gQ$ sequentially and update the representation of $u_t$ when new bonds are added to $\graph_t$. In the $k^\mathrm{th}$ step, we predict the bond type of $(u_t,q_k)$ as follows:
\begin{equation}
b_{u_t, q_k} = \softmax\left(\MLP(\vg_{u_t}^{(k-1)}, \vh_{q_k}^{(t)}, \vh_{\graph_t}, \vz_\graph)\right) 
\end{equation}
where $\vg_{u_t}^{(k-1)}$ is the new representation of $u_t$ after bonds $\set{(u_t,q_1),\cdots,(u_t,q_{k-1})}$ have been added to $\graph_t$:
\begin{equation}
\vg_{u_t}^{(k-1)} = \MLP\left(\ve(a_{u_t}), \sum_{j=1}^{k-1}\nolimits \MLP(\vh_{q_j}^{(t)}, \ve(b_{q_j,u_t})) \right) \nonumber
\end{equation}

\subsection{Training Procedure}
\label{sec:training}
Our training objective is to maximize the expected reward of generated molecules $\graph$, where the reward is an indicator of $r_i(\graph) \geq \delta_i$ for all properties $1\leq i \leq M$
\begin{equation}
\sum_\graph\nolimits \mathbb{I}\left[\bigwedge_i\nolimits r_i(\graph) \geq \delta_i\right] P(\graph) + \lambda \mathbb{H}[P(\gS)] \label{eq:formulation}
\end{equation}
We incorporate an entropy regularization term $\mathbb{H}[P(\gS)]$ to encourage the model to explore different types of rationales. 
The rationale distribution $P(\gS)$ is a categorical distribution over the rationale vocabulary. 
Let $\mathbb{I}[\graph] = \mathbb{I}\left[\bigwedge_i\nolimits r_i(\graph) \geq \delta_i\right]$. It is easy to show that the optimal $P(\gS)$ has a closed form solution:
\begin{equation}
    P(\gS_k) \propto \exp\left(\frac{1}{\lambda} \sum_\graph\nolimits \mathbb{I}[\graph] P(\graph | \gS_k)  \right) \label{eq:ratdist}
\end{equation}

The remaining question is how to train graph generator $P(\graph | \gS)$.
The generator seeks to produce molecules that are realistic and positive.
However, Eq.(\ref{eq:formulation}) itself does not take into account whether generated molecules are realistic or not. To encourage the model to generate realistic compounds, we train the graph generator in two phases:
\begin{itemize}[leftmargin=*,topsep=0pt,itemsep=0pt]
    \item \emph{Pre-training} $P(\graph | \gS)$ using real molecules.
    \item \emph{Fine-tuning} $P(\graph | \gS)$ using policy gradient with reward from property predictors.
\end{itemize}
The overall training algorithm is shown in Algorithm~\ref{alg:train}.

\begin{algorithm}[t]
\caption{Training method with $n$ property constraints.}
\label{alg:train}
\renewcommand\algorithmiccomment[1]{\hfill $\triangleright$ {#1}}
\begin{algorithmic}[1]
   \FOR{$i=1 \;\mathrm{to}\; M$}
   \STATE $V_\gS^i \leftarrow$ rationales extracted from existing molecules $\gD_i^{pos}$ positive to property $i$. \hfill  (see \S\ref{sec:rationale})
   \ENDFOR
   \STATE Construct multi-property rationales $V_\gS^{[M]}$.
   \STATE Pre-train $P(\graph | \gS)$ on the pre-training dataset $\gD^{pre}$.
   \STATE Fine-tune model $P(\graph | \gS)$ on $\gD^f$ for $L$ iterations using policy gradient.
   \STATE Compute $P(\gS)$ based on Eq.(\ref{eq:ratdist}) using fine-tuned model $P(\graph | \gS)$. 
\end{algorithmic}
\end{algorithm}

\subsubsection{Pre-training}
\label{sec:pretraining}
In addition to satisfying all the property constraints, the output of the model should constitute a realistic molecule.
For this purpose, we pre-train the graph generator on a large set of molecules from ChEMBL \citep{gaulton2017chembl}. Each training example is a pair $(\gS, \graph)$, where $\gS$ is a random connected subgraph of a molecule $\graph$ with up to $N$ atoms. The task is to learn to expand a subgraph into a full molecule. In particular, we train the generative model $P(\graph | \gS)$ to maximize the likelihood of the pre-training dataset $\gD^{pre}=\set{(\gS_i, \graph_i)}_{i=1}^n$.

\subsubsection{Fine-tuning}
After pre-training, we further fine-tune the graph generator on property-specific rationales $\gS \in \gV_\gS$ in order to maximize Eq.(\ref{eq:formulation}).  
The model is fine-tuned through multiple iterations using policy gradient \citep{sutton2000policy}. Let $P_{\theta^t}(\graph | \gS)$ be the model trained till $t^\mathrm{th}$ iteration. In each iteration, we perform the following two steps:
\begin{enumerate}[leftmargin=*,topsep=0pt,itemsep=0pt]
    \item Initialize the fine-tuning set $\gD^f=\emptyset$. For each rationale $\gS_i$, use the current model to sample $K$ molecules $\set{\graph_i^1, \cdots, \graph_i^K} \sim P_{\theta^t}(\graph | \gS_i)$. Add $(\graph_i^k, \gS_i)$ to set $\gD^f$ if $\graph_i^k$ is predicted to be positive. 
    
    \item Update the model $P_{\theta}(\graph | \gS)$ on the fine-tuning set $\gD^f$ using policy gradient method.
\end{enumerate}
After fine-tuning $P(\graph | \gS)$, we compute the rationale distribution $P(\gS)$ based on Eq.(\ref{eq:ratdist}).
\section{Experiments}
\newcommand\Tstrut{\rule{0pt}{2.3ex}}
\newcommand\Bstrut{\rule[-0.9ex]{0pt}{0pt}}

\begin{table*}[t]
\centering
\vspace{-5pt}
\caption{Results on molecule design with one or two property constraints.}
\vspace{5pt}
\begin{tabular}{lccccccccc}
\hline
\multirow{2}{*}{Method} & \multicolumn{3}{c}{ GSK3$\beta$ } & \multicolumn{3}{c}{JNK3} & \multicolumn{3}{c}{GSK3$\beta$ + JNK3} \Tstrut\Bstrut \\
\cline{2-10}
 & Success & Novelty & Diversity & Success & Novelty & Diversity & Success & Novelty & Diversity  \Tstrut\Bstrut \\
\hline
JT-VAE & 32.2\% & 11.8\% & 0.901 & 23.5\% & 2.9\% & 0.882 & 3.3\% & 7.9\% & \textbf{0.883} \Tstrut\Bstrut \\
GCPN & 42.4\% & 11.6\% & \textbf{0.904} & 32.3\% & 4.4\% & \textbf{0.884} & 3.5\% & 8.0\% & 0.874 \Tstrut\Bstrut \\
GVAE-RL & 33.2\% & 76.4\% & 0.874 & 57.7\% & 62.6\% & 0.832 & 40.7\% & 80.3\% & 0.783 \Tstrut\Bstrut \\
REINVENT & 99.3\% & \textbf{61.0\%} & 0.733 & 98.5\% & 31.6\% & 0.729 & 97.4\% & 39.7\% & 0.595 \Tstrut\Bstrut \\
\hline
RationaleRL & \textbf{100\%} & 53.4\% & 0.888 & \textbf{100\%} & \textbf{46.2\%} & 0.862 & \textbf{100\%} & \textbf{97.3\%} & 0.824 \Tstrut\Bstrut \\
\hline
\end{tabular}
\label{tab:single}
\end{table*}

\begin{table}[t]
\centering
\vspace{-5pt}
\caption{Molecule design with four property constraints. The novelty and diversity of JT-VAE, GVAE-RL and GCPN are not reported due to their low success rate.}
\vspace{5pt}
\begin{tabular}{lccc}
\hline
\multirow{2}{*}{Method} & \multicolumn{3}{c}{ GSK3$\beta$ + JNK3 + QED + SA} \Tstrut\Bstrut  \\
\cline{2-4}
 & Success & Novelty & Diversity \Tstrut\Bstrut \\
\hline
JT-VAE & 1.3\% & - & - \Tstrut\Bstrut \\
GVAE-RL & 2.1\% & - & - \Tstrut\Bstrut \\
GCPN & 4.0\% & - & - \Tstrut\Bstrut \\
REINVENT & 47.9\% & 56.1\% & 0.621 \Tstrut\Bstrut \\
\hline
RationaleRL & \textbf{74.8\%} & \textbf{56.8\%} & \textbf{0.701}  \Tstrut\Bstrut \\
\hline
\end{tabular}
\label{tab:triple}
\end{table}

\begin{figure*}[t]
    \centering
    \includegraphics[width=\textwidth]{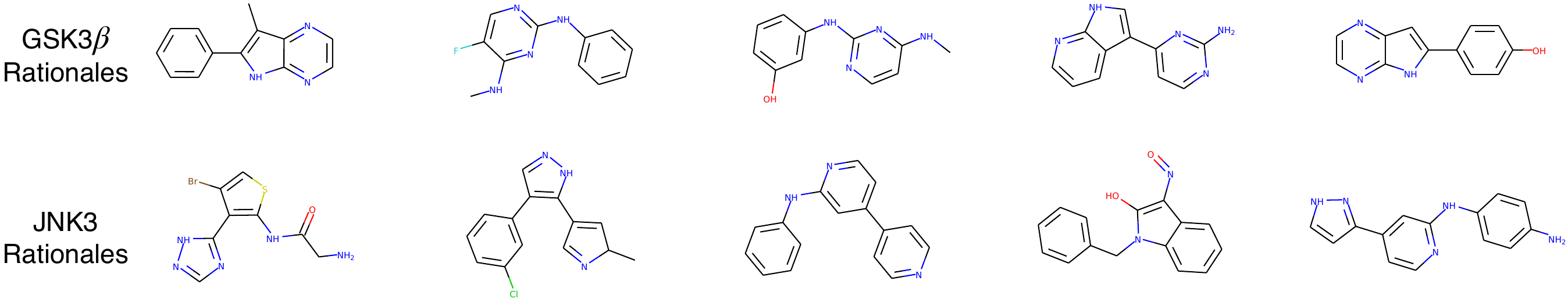}
    \caption{Sample rationales of GSK3$\beta$ (top) and JNK3 (bottom).}
    \label{fig:samplerats}
\end{figure*}

\begin{figure*}[t]
    \centering
    \hspace{-10pt}
    \begin{subfigure}{0.3\textwidth}
        \centering
        \includegraphics[width=\textwidth]{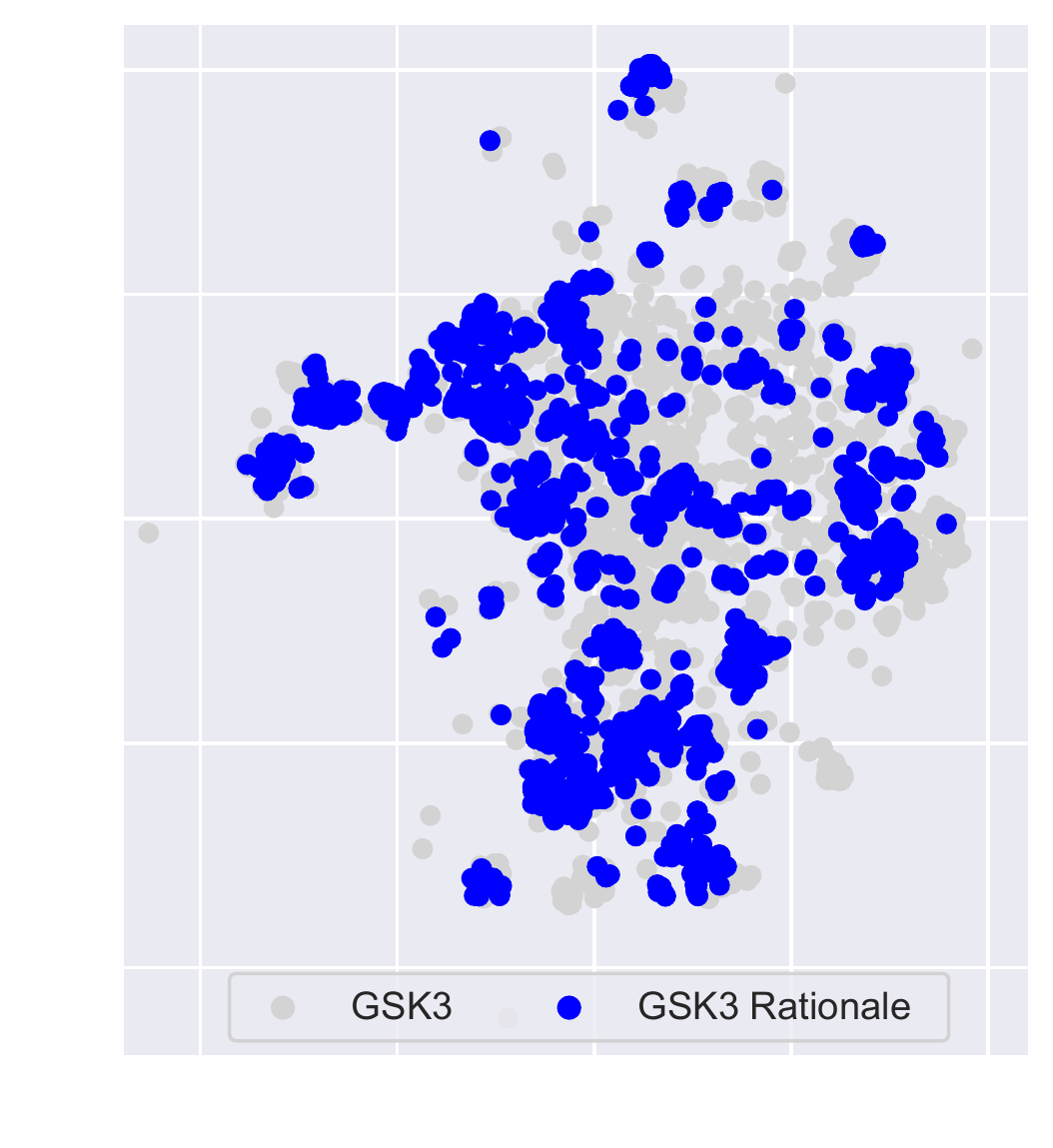}
    \end{subfigure}
    ~
    \begin{subfigure}{0.3\textwidth}
        \centering
        \includegraphics[width=\textwidth]{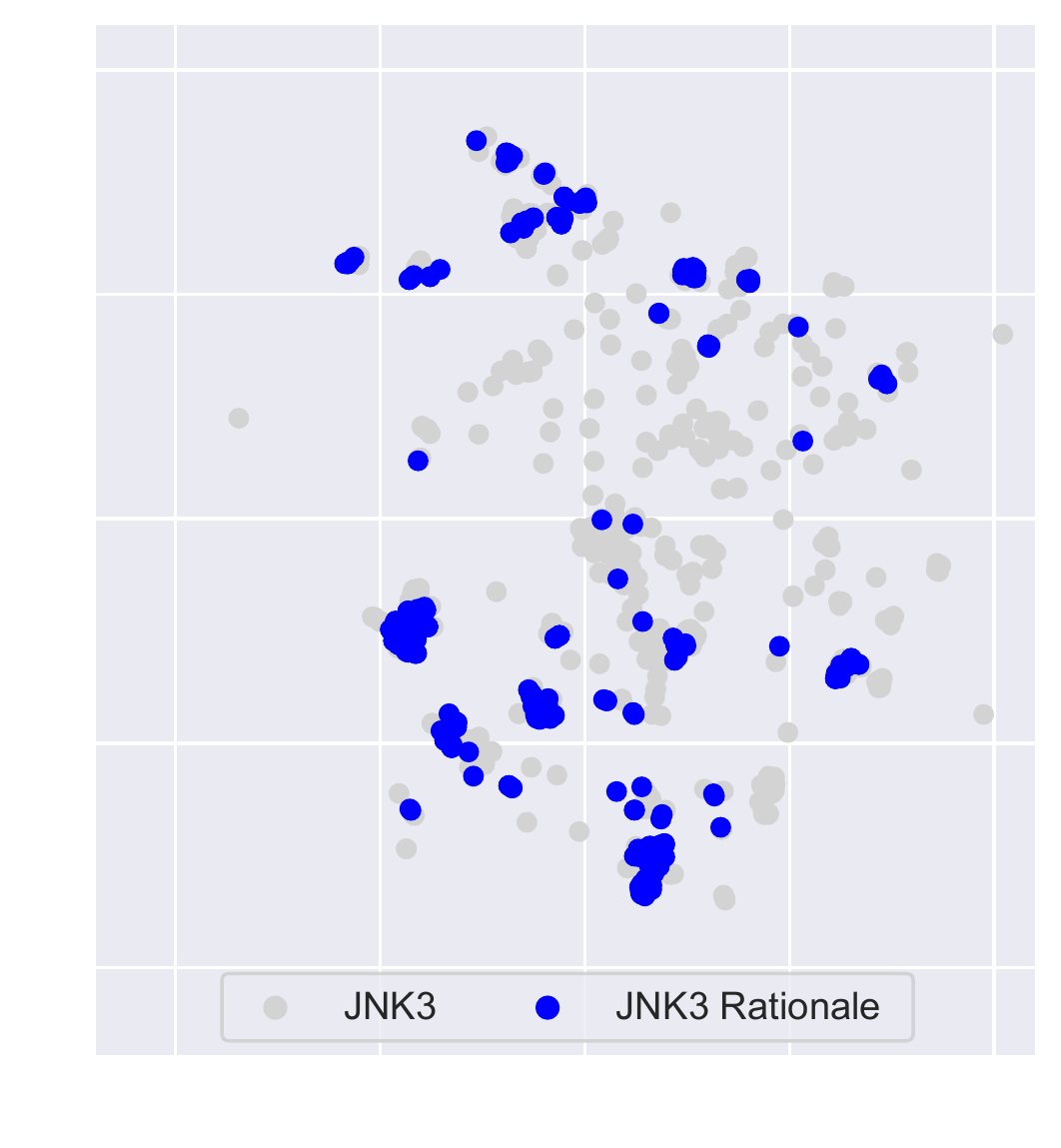}
    \end{subfigure}
    ~
    \begin{subfigure}{0.29\textwidth}
        \centering
        \includegraphics[width=\textwidth]{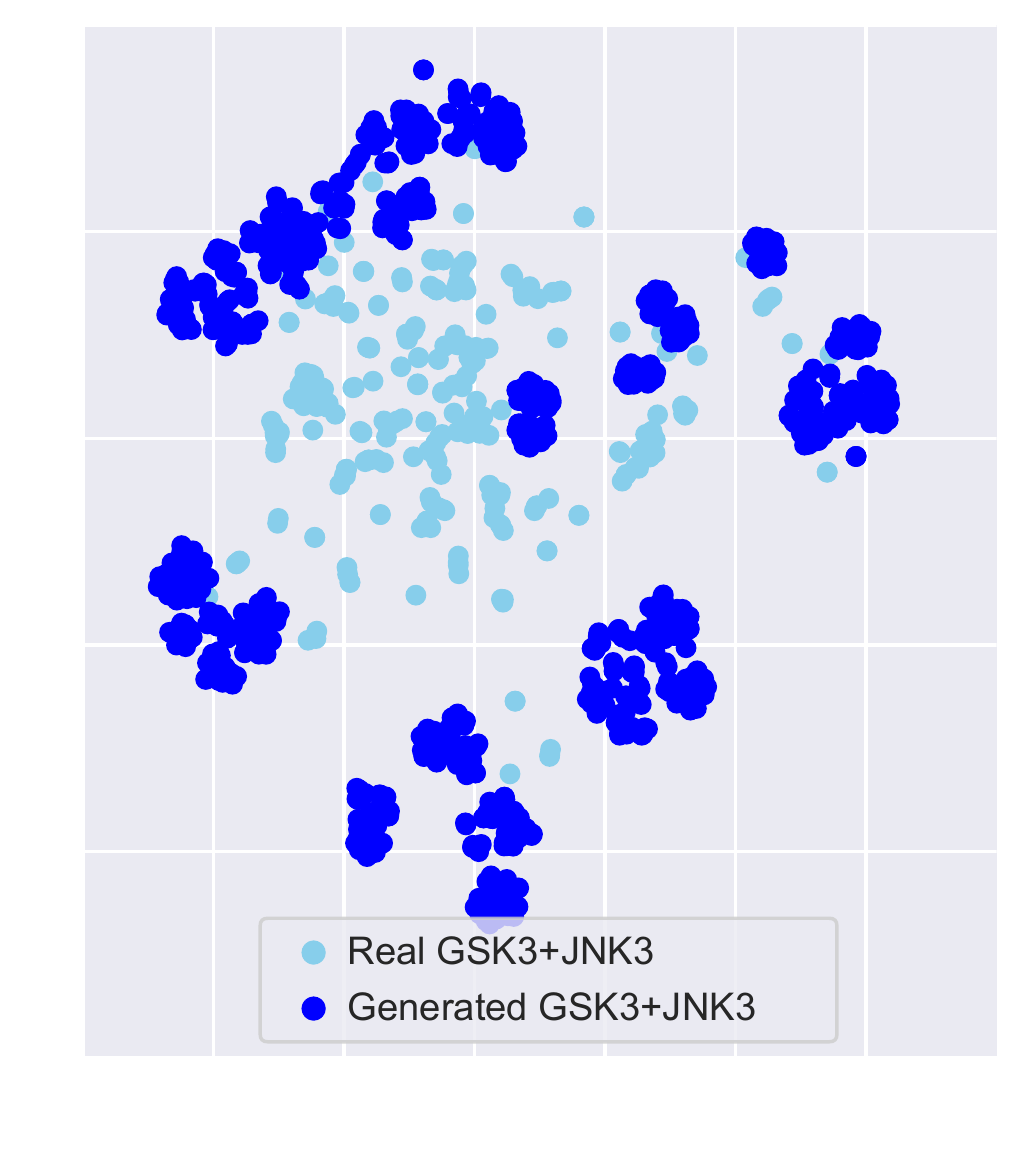}
    \end{subfigure}
    \vspace{-8pt}
    \caption{\textbf{Left \& middle}: t-SNE plot of the extracted rationales for GSK3$\beta$ and JNK3. For both properties, rationales mostly covers the chemical space populated by existing positive molecules. \textbf{Right}: t-SNE plot of generated GSK3$\beta$+JNK3 dual inhibitors.}
    \label{fig:vis1}
\end{figure*}

\begin{figure*}[t]
    \centering
    \includegraphics[width=\textwidth]{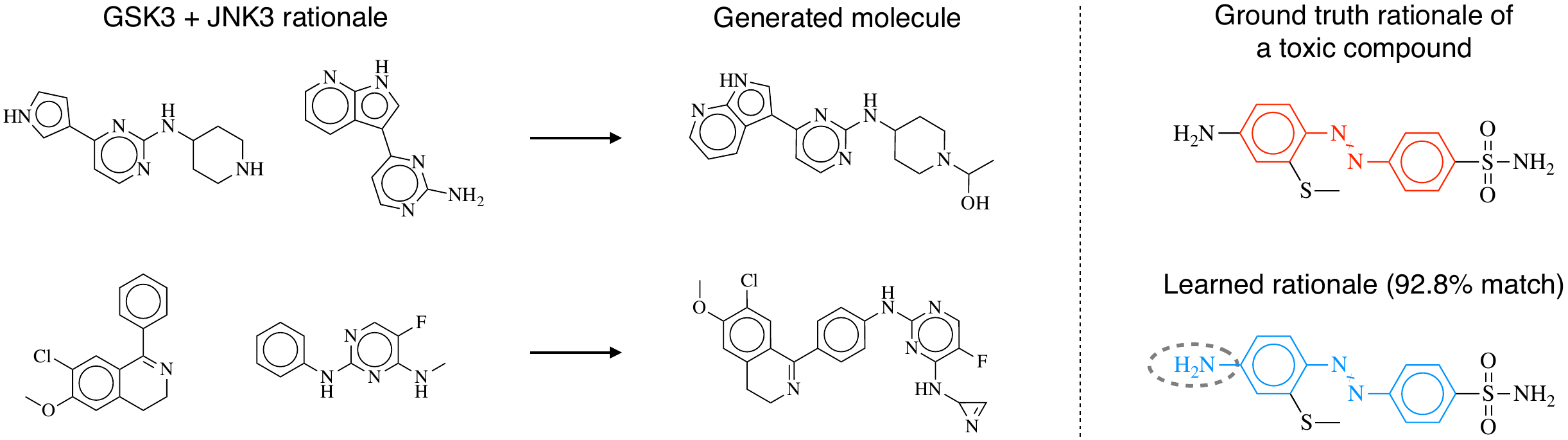}
    \caption{\textbf{Left}: Examples of molecules generated in the GSK3$\beta$+JNK3+QED+SA task. The model learns to combine two disjoint rationale graphs into a complete molecule. \textbf{Right}: Example structural alerts in the toxicity dataset. The ground truth rationale (Azobenzene) is highlighted in red. Our learned rationale almost matches the ground truth (error highlighted in dashed circle).}
    \label{fig:vis2}
\end{figure*}

We evaluate our method (RationaleRL) on molecule design tasks under various combination of property constraints. In our experiments, we consider the following two properties:
\begin{itemize}[leftmargin=*,topsep=0pt,itemsep=0pt]
    \item GNK3$\beta$: Inhibition against glycogen synthase kinase-3 beta. The GNK3$\beta$ prediction model is trained on the dataset from \citet{li2018multi}, which contains 2665 positives and 50K negative compounds.
    \item JNK3: Inhibition against c-Jun N-terminal kinase-3. The JNK3 prediction model is also trained on the dataset from \citet{li2018multi} with 740 positives and 50K negatives.
\end{itemize}
Following \citet{li2018multi}, the property prediction model is a random forest using Morgan fingerprint features~\citep{rogers2010extended}. We set the positive threshold $\delta_i=0.5$. For each property, we split its property dataset into 80\%, 10\% and 10\% for training, validation and testing. The test AUROC score is 0.86 for both GSK3$\beta$ and JNK3.

\textbf{Multi-property Constraints } We also consider combinations of property constraints that are biologically relevant:
\begin{itemize}[leftmargin=*,topsep=0pt,itemsep=0pt]
    \item GNK3$\beta$ + JNK3: Jointly inhibiting JNK3 and GSK-3$\beta$ may provide potential benefit for the treatment of Alzheimer’s disease~\citep{li2018multi}. There exist 316 dual inhibitors already available in the dataset.
    \item GNK3$\beta$ + JNK3 + QED + SA: We further require the generated dual inhibitors to be drug like and synthetically accessible (i.e., easy to be synthesized). These two properties are quantified by QED~\cite{bickerton2012quantifying} and synthetic accessibility (SA) score~\cite{ertl2009estimation}. In particular, we require $\mathrm{QED} \geq 0.6$ and $\mathrm{SA} \leq 4.0$.
\end{itemize}

\textbf{Evaluation Metrics } Our evaluation effort measures various aspects of molecule design. For each method, we generate $n=5000$ molecules and compute the following metrics:
\begin{itemize}[leftmargin=*,topsep=0pt,itemsep=0pt]
    \item \textbf{Success}: The fraction of sampled molecules predicted to be positive (i.e., satisfying all property constraints). A good model should have a high success rate. Following previous work~\citep{olivecrona2017molecular,you2018graph}, we only consider the success rate under property predictors, as it is hard to obtain real property measurements for GSK3$\beta$ and JNK3. 
    \item \textbf{Diversity}: It is also important for a model generate diverse range of positive molecules. To this end, we measure the diversity of generated positive compounds by computing their pairwise molecular distance $\mathrm{sim}(X,Y)$, which is defined as the Tanimoto distance over Morgan fingerprints of two molecules.
    $$
    \mathrm{Diversity} = 1 - \frac{2}{n(n-1)} \sum_{X,Y}\nolimits \mathrm{sim}(X,Y)
    $$
    \item \textbf{Novelty}: Crucially, a good model should discover novel positive compounds. In this regard, for each generated positive compound $\graph$, we find its nearest neighbor $\graph_{\mathrm{SNN}}$ from positive molecules in the training set. We define the novelty as the fraction of molecules with nearest neighbor similarity lower than 0.4~\citep{olivecrona2017molecular}:
    $$
    \mathrm{Novelty} = \frac{1}{n} \sum_{\graph}\nolimits \mathbf{1} \left[ \mathrm{sim}(\graph, \graph_{\mathrm{SNN}}) < 0.4 \right]
    $$
\end{itemize}

\textbf{Baselines } We compare our method against the following state-of-the-art generation methods for molecule design:
\begin{itemize}[leftmargin=*,topsep=0pt,itemsep=0pt]
    \item JT-VAE~\citep{jin2018junction} is a generative model which generate molecules based on substructures such as rings and bonds. The model contains auxiliary property predictors over the VAE latent space. During testing, we perform multi-objective optimization in the latent space using the gradient from the property predictors.
    \item REINVENT \citep{olivecrona2017molecular} is a RL model generating molecules based on their SMILES strings \citep{weininger1988smiles}. To generate realistic molecules, their model is pre-trained over one million molecules from ChEMBL and then finetuned under property reward.
    \item GCPN~\citep{you2018graph} is a RL model which generates molecular graphs atom by atom. It uses GAN~\citep{goodfellow2014generative} to help generate realistic molecules.
    \item GVAE-RL is a graph variational autoencoder which generates molecules atom by atom. The graph VAE architecture is the same as our model, but it generates molecules from scratch without using rationales. The model is pre-trained on the same ChEMBL dataset and fine-tuned for each property using policy gradient. This is an ablation study to show the importance of using rationales.
\end{itemize}

\textbf{Rationales } Details of the rationales used in our model:
\begin{itemize}[leftmargin=*,topsep=0pt,itemsep=0pt]
    \item GSK3$\beta$, JNK3: For single properties, the rationale size is required to be less than 20 atoms. For each positive molecule, we run 20 iteration of MCTS with $c_{puct}=10$. There are 6299 and 417 rationales for GSK3$\beta$ and JNK3.
    \item GSK3$\beta$ + JNK3: We construct two-property rationales by merging the single-property rationales (see Figure~\ref{fig:merge}). For computational efficiency, we only consider rationales $\gS_k$ receiving the highest $P(\gS_k)$ within each property. This gives us two rationale subsets with $|V_\gS^\mathrm{GSK3\beta}|=238$ and $|V_\gS^\mathrm{JNK3}|=39$. In total, there are 9719 two-property rationales merged from $V_\gS^\mathrm{GSK3\beta}$ and $V_\gS^\mathrm{JNK3}$.
    \item GSK3$\beta$ + JNK3 + QED + SA: We construct four-property rationales by selecting the two-property rationales whose $\mathrm{QED} \geq 0.6$ and $\mathrm{SA} \leq 4.0$. There are in total 61 different four-property rationales.
\end{itemize}

\textbf{Model Setup } We pre-train all the models on the same ChEMBL dataset, which contains 1.02 million training examples. On the four-property generation task, our model is fine-tuned for $L=50$ iterations, with each rationale being expanded for $K=200$ times.

\subsection{Results}

The results are reported in Table~\ref{tab:single} and \ref{tab:triple}. On the single-property generation task, both REINVENT and RationaleRL demonstrate nearly perfect success rate since there is only one constraint.
On the two-property generation task, our model achieves 100\% success rate while maintaining 97.3\% novelty and 0.824 diversity score, which is much higher than REINVENT. Meanwhile, GCPN has a low success rate (3.5\%) due to reward sparsity.

Table~\ref{tab:triple} shows the results on the four-property generation task, which is the most challenging. The difference between our model the baselines become significantly larger. In fact, GVAE-RL and GCPN completely fail in this task due to reward sparsity. Our model outperforms REINVENT with a wide margin (success rate: 74.8\% versus 47.9\%; diversity: 0.701 versus 0.621).

\textbf{Ablation study } In all tasks, our method significantly outperforms GVAE-RL, which has the same generative architecture but does \emph{not} utilize rationales and generates molecules from scratch. Thus we conclude the importance of rationales for multi-property molecular design.

\textbf{Visualization } We further provide visualizations to help understand our model. In Figure~\ref{fig:vis1}, we plotted a t-SNE~\citep{maaten2008visualizing} plot of the extracted rationales for GSK3$\beta$ and JNK3. For both properties, rationales mostly cover the chemical space populated by existing positive molecules. The generated GSK3$\beta$+JNK3 dual inhibitors are distributionally close to the true dual inhibitors in the training set. In Figure~\ref{fig:samplerats}, we show samples of rationales extracted by MCTS. In Figure~\ref{fig:vis2}, we show examples of generated molecules that satisfy all the four constraints (GSK3$\beta$+JNK3+QED+SA).

\subsection{Property Predictor Applicability}

Since the reported success rate is based on property predictors, it is possible for the generated molecules to exploit flaws of the property predictors. In particular, the predicted properties may be unreliable when the distribution of generated compounds is very different from the distribution of molecules used to train the property predictors.

This issue is alleviated by our framework because our generated compounds are built from rationales extracted from true positive compounds used to train the property predictors. Therefore, our generated compounds are closer to the true compounds than compounds generated from scratch. To show this, we adopt Frechet ChemNet Distance (FCD)\footnote{The FCD score is analogous to the Frechet Inception Distance~\cite{heusel2017gans} used in image synthesis.} to measure distributional discrepancy between generated molecules and positive compounds in the training set~\cite{preuer2018frechet}.
As shown in Table~\ref{tab:applicability}, the FCD of RationaleRL is much lower than REINVENT, which means our generated compounds are distributionally closer to the true compounds. As a result, the predicted properties are more reliable for molecules generated by our model. 

\begin{table}[t]
\centering
\vspace{-5pt}
\caption{Frechet ChemNet Distance (FCD) between generated compounds and true positive molecules in the training set.}
\vspace{5pt}
\begin{tabular}{lccc}
\hline
& GSK3$\beta$ & JNK3  & GSK3$\beta$+JNK3\Tstrut\Bstrut \\
\hline
REINVENT & 28.63 & 25.16 & 36.75 \Tstrut\Bstrut \\
RationaleRL & \textbf{6.88} & \textbf{9.68} & \textbf{25.52} \Tstrut\Bstrut \\
\hline
\end{tabular}
\label{tab:applicability}
\end{table}

\subsection{Faithfulness of Rationales}

\begin{table}[t]
\centering
\vspace{-5pt}
\caption{Rationale accuracy on the toxicity dataset. Our rationales are more faithful to the property of interest.}
\vspace{5pt}
\begin{tabular}{lcc}
\hline
Method & Partial Match & Exact Match \Tstrut\Bstrut \\
\hline
Integrated Gradient & 0.857 & 39.4\% \Tstrut\Bstrut \\
MCTS Rationale & \textbf{0.861} & \textbf{46.0\%}  \Tstrut\Bstrut \\
\hline
\end{tabular}
\label{tab:rationale}
\end{table}

While our rationales are mainly extracted for generation, it is also important for them to be chemically relevant. In other words, the extracted rationales should accurately explain the property of interest. As there is no ground truth rationales available for JNK3 and GSK3$\beta$, we turn to an auxiliary toxicity dataset for evaluating rationale quality.

\textbf{Data } The toxicity dataset contains 125K molecules randomly selected from ChEMBL. Each molecule is labeled as toxic if it contains structural alerts~\citep{sushko2012toxalerts} --- chemical substructures that is correlated with human or environmental hazards (see Figure~\ref{fig:vis2}).\footnote{Structural alerts used in our paper are from surechembl.org/ knowledgebase/169485-non-medchem-friendly-smarts}
Under this setup, the structural alerts are ground truth rationales and we evaluate how often the extracted rationales match them.
The dataset is split into 105K for training and 20K for testing. In total, there are 26.5K toxic molecules and 164 types of structural alerts. We train a graph convolutional network~\citep{yang2019analyzing} to predict toxicity, which achieves 0.99 AUROC score on the test set.

\textbf{Results } We compare our MCTS based rationale extraction with integrated gradient~\citep{sundararajan2017axiomatic}, which has been applied to explain property prediction models~\citep{mccloskey2019using}. We report two metrics: \emph{partial match} AUC (attribution AUC metric used in \citet{mccloskey2019using}) and \emph{exact match} accuracy which measures how often a rationale graph exactly matches the true rationale in the molecule. 
As shown in Table~\ref{tab:rationale}, our method significantly outperforms the baseline in terms of exact matching. The extracted rationales has decent overlap with true rationales, with 0.86 partial match on average. Therefore, our model is capable of finding rationales faithful to the properties.

\section{Conclusion}
In this paper, we developed a rationale based generative model for molecular design. Our model generates molecules in two phases: 1) identifying rationales whose presence indicate strong positive signals for each property; 2) expanding rationale graphs into molecules using graph generative models and fine-tuning it towards desired combination of properties. Our model demonstrates strong improvement over prior reinforcement learning methods in various tasks.

\section*{Acknowledgements}
We would like to thank Patrick Walters, Jiaming Luo, Adam Yala, Benson Chen, Adam Fisch, Yujia Bao and Rachel Wu for their insightful comments. We also thank the anonymous reviewers for their helpful feedback. This work was supported by MLPDS, DARPA AMD project and J-Clinic.


\bibliography{main}
\bibliographystyle{icml2020}

\end{document}